\documentclass[10pt,twocolumn,letterpaper]{article}

\usepackage{cvpr}              
\usepackage{booktabs}
\usepackage{array}
\usepackage{multirow}
\usepackage{color}
\usepackage{colortbl}

\definecolor{liu}{RGB}{255,105,97}

\definecolor{cvprblue}{rgb}{0.21,0.49,0.74}
\usepackage[pagebackref,breaklinks,colorlinks,allcolors=cvprblue]{hyperref}
\usepackage{balance}

\def\paperID{17083} 
\def\confName{CVPR}
\def\confYear{2026}

\newcommand{\NAME}{{PhysAlign}\xspace}

\title{\NAME:
Physics-Coherent Image-to-Video Generation through Feature and 3D Representation Alignment}

\author{Zhexiao Xiong$^{1}$ \quad Yizhi Song$^{2}$ \quad Liu He$^{2}$ \quad Wei Xiong$^{3}$ \quad Yu Yuan$^{2}$ \quad Feng Qiao$^{1}$ \quad Nathan Jacobs$^{1}$ \\ 
{\normalsize $^1$ Washington University in St. Louis} \quad
{\normalsize $^2$ Purdue University} \quad
{\normalsize $^3$ NVIDIA} \quad \\
}

\begin{document}

\twocolumn[{
\renewcommand\twocolumn[1][]{#1}
\maketitle
\vspace{-25pt}
\begin{center}
    \vspace{-5pt}
    \includegraphics[width=\linewidth]{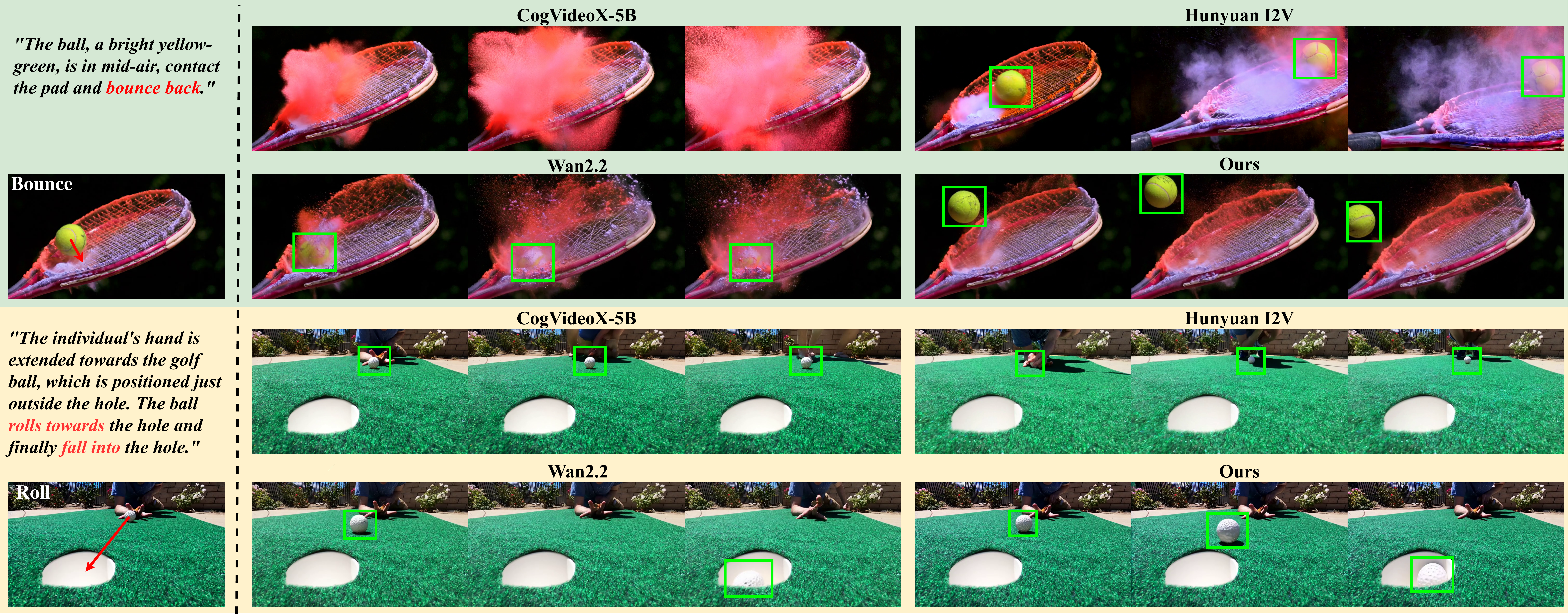}
    \vspace{-15pt}
    \captionsetup{hypcap=false}
    \captionof{figure}{Comparisons of image-to-video (I2V) generation results on physics-involved scenarios. \textcolor{red}{Red arrows} indicate the initial motion directions of the objects, while \textcolor{green}{green rectangles} highlight key subjects that reveal underlying physical behaviors and dynamics. We introduce \textbf{\NAME}, an I2V framework that effectively infuses physical knowledge and 3D geometric priors into existing video generation models. \NAME significantly enhances physical coherence and 3D perceptual fidelity, generating videos that most faithfully conform to real-world physical laws.}
    \label{fig:teaser}
\end{center}
}]

\begin{abstract}
  Video Diffusion Models (VDMs) offer a promising approach for simulating dynamic scenes and environments, with broad applications in robotics and media generation. However, existing models often generate temporally incoherent content that violates basic physical intuition, significantly limiting their practical applicability. We propose \textbf{PhysAlign}, an efficient framework for physics-coherent image-to-video (I2V) generation that explicitly addresses this limitation. To overcome the critical scarcity of physics-annotated videos, we first construct a fully controllable synthetic data generation pipeline based on rigid-body simulation, yielding a highly-curated dataset with accurate, fine-grained physics and 3D annotations. Leveraging this data, PhysAlign constructs a unified physical latent space by coupling explicit 3D geometry constraints with a Gram-based spatio-temporal relational alignment that extracts kinematic priors from video foundation models. Extensive experiments demonstrate that PhysAlign significantly outperforms existing VDMs on tasks requiring complex physical reasoning and temporal stability, without compromising zero-shot visual quality. PhysAlign shows the potential to bridge the gap between raw visual synthesis and rigid-body kinematics, establishing a practical paradigm for genuinely physics-grounded video generation. The project page is available at \url{https://physalign.github.io/PhysAlign}.
\end{abstract}

\section{Introduction}

Recent advances in video generation have enabled impressive visual quality and have shown strong potential for content creation and simulating dynamic environments. However, current VDMs still struggle with physical coherence, often producing motions or object interactions that violate basic physical laws~\citep{Kang_2024_Farvideogenerationworld, Bansal_2024_Videophy, Zhang_2025_Morpheus, Bansal_2025_Videophy2, yuan2025likephys}. Such inconsistencies limit their reliability for applications requiring strict physical grounding and motivate the need for models that generate dynamics aligned with physical reality and human perception.

To clarify what constitutes physical coherence in image-to-video (I2V) generation, we focus on two key dimensions:
\begin{itemize}
\item \textbf{General physical laws.} Generated motions should obey fundamental physics, e.g., consistent gravitational acceleration, physically plausible collisions, and momentum-conserving trajectories.
\item \textbf{3D perceptual fidelity.} The generated video should respect 3D spatial structure and perspective, including correct occlusion ordering and size changes as objects move relative to the camera.
\end{itemize}

Current VDMs are typically trained on large-scale video datasets and thus learn physical coherence only implicitly. Yet explicit physical cues—such as depth, force, mass, and object state~\citep{gillman2025force,liu2025idcnetguidedvideodiffusion,liang2025UniFuture,hassan2025gem}—provide strong supervision for physics-aware I2V generation but are rarely available in real video corpora. To overcome this limitation, we build a fully controllable synthetic data generation pipeline that simulates abstract scenes with explicit physical representations, including depth, mass, force, and initial motion angle. This enables us to generate training videos with accurate 3D and physical annotations, allowing the model to learn fine-grained dynamics and causal motion patterns during training.

To effectively internalize these dynamics into foundation models, we introduce \textbf{PhysAlign}, a unified representation alignment framework. Rather than training from scratch, PhysAlign operates as an efficient adapter for existing DiT-based~\cite{peebles2023scalable} backbones. It bridges the gap between raw visual semantics and physical rules by grounding the generator's latent space in two complementary domains: explicit 3D geometry (capturing spatial perspective) and temporally-aware foundation embeddings (extracting kinematic causality). This dual grounding enforces a deep coherence between the generated visuals and underlying physical dynamics. Extensive quantitative and qualitative experiments demonstrate that fine-tuning on merely 3K synthetic videos significantly boosts physical coherence and exhibits robust generalization to complex real-world data, all without compromising the base model's zero-shot general visual quality. 

Our contributions are summarized as follows:
\begin{itemize}
    \item \textbf{Methodology:} We propose \textbf{PhysAlign}, an efficient adapter framework for physics-coherent I2V generation. Instead of rigid token matching, we introduce a \textbf{Gram-based spatio-temporal relational alignment} to extract kinematic priors from video foundation models (e.g., V-JEPA2), and synergistically couple it with explicit 3D geometry constraints to build a unified physical latent space.

    \item \textbf{Data \& Open-Source Resource:} We construct a fully controllable synthetic video data generation pipeline based on rigid-body simulation, providing paired RGB sequences, dense 3D representations, and granular physics annotations. To facilitate future research in physics-aware video generation, we will publicly release the complete data generation engine and dataset.
    
    \item \textbf{Data Efficiency \& Generalization:} We demonstrate remarkable data efficiency. By fine-tuning on merely 3K synthetic clips, PhysAlign robustly internalizes physical laws and generalizes to complex real-world dynamics. It significantly boosts physical coherence across multiple benchmarks without compromising the base model's zero-shot general visual quality.
\end{itemize}
\section{Related Work}

\subsection{Physics-aware Video Generation}
\label{sec:physics}

Recent works have explored incorporating physical knowledge into generative
video models to improve plausibility and controllability. Rather than grouping
methods by whether physics precedes or follows generation, we categorize them
by \emph{where} physical knowledge is injected within the generative pipeline:
\emph{input space}, \emph{latent space}, or \emph{output space}.

\noindent
\textbf{Physics in the Input Space.}
These approaches introduce physics before generation, typically by producing physically plausible trajectories or scene states using external simulation tools such as MPM \cite{Stomakhin_2013_MPM}.
The simulated dynamics serve as conditioning inputs to a video generator. Representative methods include PhysGen3D, Phys4DGen, PhysGaussian, PhysMotion, PhysDreamer, AutoVFX, SeeU and PhysCtrl \cite{Chen_2025_Physgen3d, Liu_2024_Physgen, Lin_2024_Phys4dgen, Xie_2024_Physgaussian, Tan_2024_Physmotion, Zhang_2024_Physdreamer, Hsu_2024_Autovfx, Yuan_2025_SeeU,wang2025physctrl}.
While this strategy offers explicit controllability, it requires users to manually design scenario-specific physical rules and does not generalize across environments or physical regimes.

\noindent
\textbf{Physics in the Latent Space.}
A growing set of techniques embed physics directly into the latent
representations or motion structures of generative models.
Examples include physics-guided latent embeddings, or structured dynamic primitives that impose smoothness, consistency, or conservation-like behaviors. Methods such as PhysDiff,  PhysAnimator, NewtonGen, WonderPlay, MotionMode, and VLIPP follow this direction \cite{Yuan_2023_Physdiff,  Xie_2025_Physanimator, yuan2025newtongen, li2025wonderplay, Pandey_2025_MotionMode, Yang_2025_VLIPP, Yuan_2026_Physalign}.
This family of approaches allows the generative model itself to maintain physically plausible dynamics rather than relying on precomputed simulations.

\noindent
\textbf{Physics in the Output Space.}
Another line of work enforces physics \emph{after} generation, typically through external evaluators such as large language models or vision–language models that assess physical consistency.
Examples include GPT4Motion, PhyT2V, PhyWorldBench, and related
self-refinement pipelines \cite{Lv_2024_Gpt4motion, Xue_2025_PhyT2V, gu2025phyworldbench}.
These models refine or rank generated videos based on heuristic or learned physical judgments. However, because physical reasoning is learned implicitly from data, their robustness deteriorates sharply under physically challenging or out-of-distribution scenarios.

\vspace{-10pt}
\subsection{Image-to-Video Generation}
Early video diffusion works primarily relied on text prompts, whereas conditioning on an image provides a more direct way to preserve subject appearance and spatial structure in controllable video generation. Early methods such as AnimateDiff~\citep{guo2023animatediff} inject motion modules into text-to-image diffusion to animate static content. Subsequent approaches like VideoCrafter~\citep{chen2023videocrafter1,chen2024videocrafter2} and DynamicCrafter~\citep{xing2023dynamicrafter} encode the reference image into a text-aligned latent space via a query transformer, enabling the video model to better leverage visual cues. Recent DiT-based architectures such as CogVideoX~\citep{yang2024cogvideox}, LTX-Video~\citep{HaCohen2024LTXVideo}, HunyuanVideo~\citep{motamed2025generative}, and Wan2.1/2.2~\citep{wan2025}—treat the reference image as an additional conditioning channel concatenated with noisy latents, achieving stronger spatial fidelity and scalability. Despite these advances, current I2V models mainly rely on RGB cues and lack explicit 3D or physics modeling, often producing motions that violate basic physical laws. In this work, we use Wan2.2-14B as our base model and enhance it with physics and geometry aware representation alignment to improve physical coherence.

\vspace{-10pt}
\section{Preliminary: Representation Alignment}
Representation alignment (REPA)~\citep{yu2024representation,zhang2025videorepa,leng2025repa,tian2025u,yao2025reconstruction} is a common regularization method that effectively accelerates the convergence process of diffusion models. It aligns the encoded representations of a noisy input in diffusion models with representations from other pretrained visual encoders. Specifically, given an image $x \in \mathbb{R}^{H \times W \times 3}$, semantically rich representation $\mathbf{E}_*$ is extracted from the encoder $f$ of a pre-trained vision foundation model, and the feature $\mathbf{E}_t$ of the diffusion model is obtained from the noisy input $x_t$, represented as $\mathbf{E}_t = g(x_t)$. The objective of the token-wise alignment can then be defined as:
\begin{equation}
\mathcal{L}_{\mathrm{REPA}}=-\mathbb{E}\left[\frac{1}{N} \sum_{n=1}^N \operatorname{sim}\left(\mathbf{E}_*^{[\mathbf{n}]}, \mathbf{W}\left(\mathbf{E}_t^{[\mathbf{n}]}\right)\right)\right]
\end{equation}
where \texttt{sim}(·,·) denotes a similarity metric (e.g., cosine similarity), and $\mathbf{W}$ denotes a trainable MLP layer for dimension matching. By minimizing the semantic gaps between $\mathbf{E}_t$ and $\mathbf{E}_*$, this method improves both training efficiency and generation quality when applied to diffusion and flow-based transformers such as DiT.

\begin{figure}[t!]
    \centering
    \includegraphics[width=\linewidth]{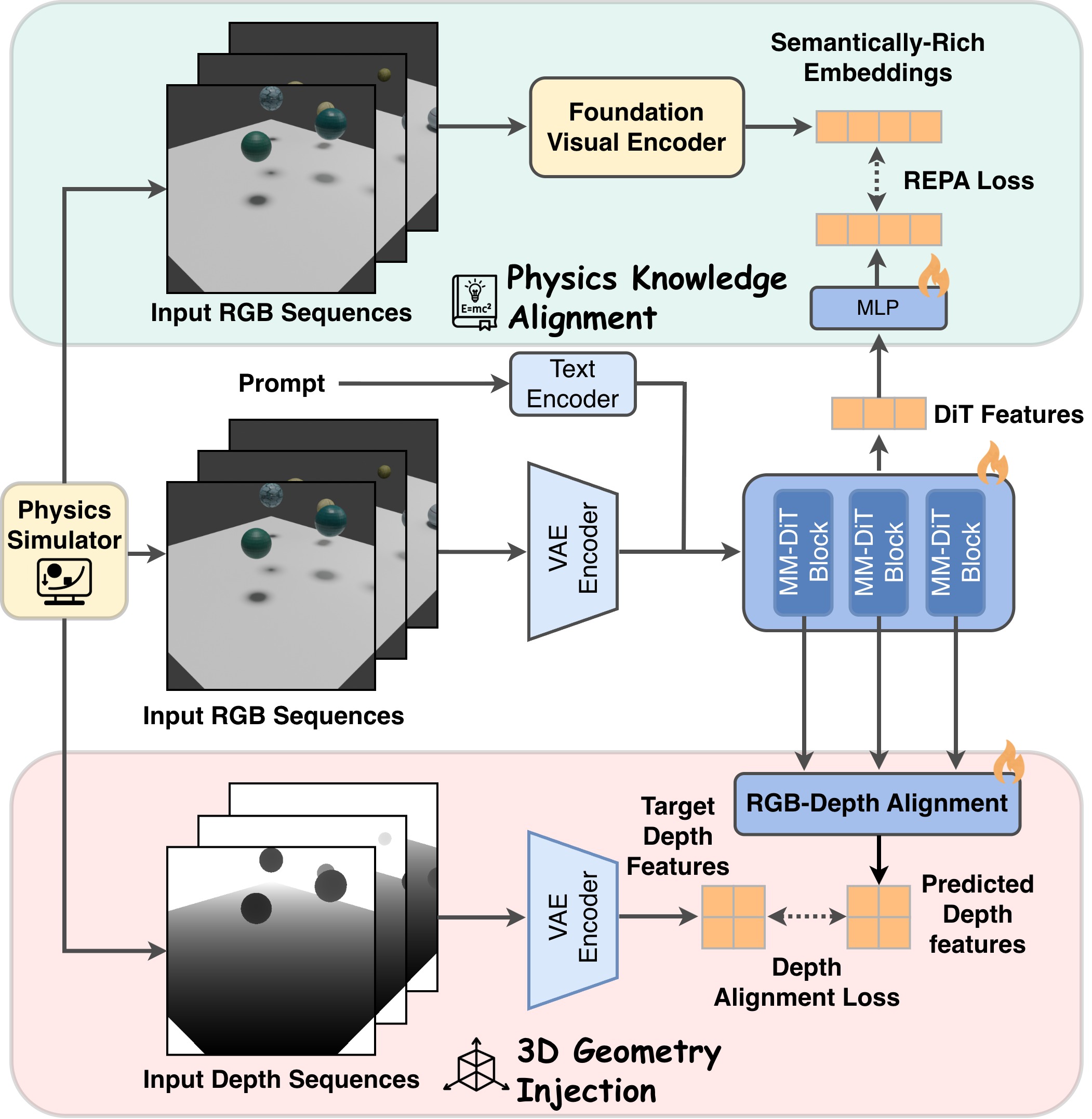}
    \caption{\textbf{\NAME} framework. Our data generation pipeline leverages physical simulator (i.e. Blender) to generate synthetic videos with 3D physical ground truth. Our method aligns the DiT~\cite{peebles2023scalable} latent features with both (i) physical knowledge feature by V-JEPA2~\citep{assran2025v}, and (ii) 3D geometric feature encoded from synthetic ground truth (e.g., depth). This unified alignment internalizes both physical laws and visual fidelity for I2V generation task.} 
    \label{fig:pipeline}
\end{figure}

\section{Method}
Given a text prompt and a single reference image, we aim to synthesize a video that is (i) physics-coherent with real-world dynamics and (ii) visually consistent in 3D space.
Our key insights are:
\begin{itemize}
    \item Better video understanding improves the physical plausibility of the videos generated by VDMs \cite{koh2023generating,liao2025imagegen,xing2024dynamicrafter,yang2025vlipp}, which can be obtained through foundational visual models;
    \item Dense 3D geometry representations enhance the motion coherence and 3D visual consistency (e.g., with correct occlusion and perspective change) of the generated videos \cite{chefer2025videojam,xie2025physanimator};
    \item The scarcity of dense 3D labels in the current video datasets hinders the efforts to inject 3D priors into VDMs \cite{nguyen2025physix}; so a controllable physics simulator can efficiently provide accurate and rich 3D clues, resolving the data bottleneck.
\end{itemize}

Motivated by the above insights, we operationalize physics coherence through two complementary signals: a \emph{physical knowledge} signal (see Sec.~\ref{sec:repa_phys}) that transfers the physics understanding of the scene into the generator, and a \emph{3D geometry} signal (see Sec.~\ref{sec:3d_inject} that ensures the consistency and realism in terms of geometry and depth. In addition, we design a physics simulator with fine-grained control (in Sec.~\ref{sec:simulator}), which generates large-scale dense 3D labels for the RGB inputs. The overall pipeline is illustrated in Fig.~\ref{fig:pipeline}.

\subsection{Physics Simulator for Data Generation}
\label{sec:simulator}

To generate physically coherent video data, we design a fine-grained and fully controllable physics simulator that serves as the foundation of our synthetic data pipeline. We build our physics engine on top of Blender, and generate rigid-body dynamics with low-level control. This simulator allows us to specify physically accurate parameters derived from real-world objects.

Using this simulator, we configure diverse physical parameters for each subject, including mass, size and elastic coefficients. These parameters are sampled from realistic ranges to emulate objects under different physical conditions. For training data, each scene contains 3–7 objects with physically grounded initial states. Specifically, we assign each object an initial horizontal velocity generated from a randomly sampled force direction (uniformly from 0°–360°), along with a random drop height. This design ensures natural variation in motion trajectories and introduces a probability of object collisions. As a result, the generated videos capture collisions and motion patterns that are consistent with real-world rigid-body physics.

\noindent \textbf{Multi-Modal Rendering.} Leveraging blender's ability to simulate accurate training data, the multi-modal generation consists of: (1) RGB frames capturing photorealistic appearance with ray-traced lighting, (2) depth maps encoding metric distance from camera. The camera orientation is randomly sampled within a predefined range. By generating data with geometry, we can ensure 3D geometry-conditioned generation. For the text annotations, we fuse all the physics parameters(force, mass, height, etc) into the prompt as conditions.

\subsection{Physical Knowledge Injection via Video Representation Alignment}
\label{sec:repa_phys}

We begin by introducing physical knowledge and scene understanding into the VDM. The physical plausibility of a generated video depends critically on understanding not just who moves (object identity), but the causal relationships of how things move over time (kinematics, inertia, and collisions). While static image encoders provide rich spatial semantics, they fundamentally lack temporal context. Therefore, we utilize a temporally-aware video foundation model (V-JEPA2) as our teacher. By encoding video clips jointly, V-JEPA2 produces motion-aware patch embeddings that inherently capture temporal evolution and dynamic object interactions. We inject these physical cues into the generator by aligning the \emph{spatio-temporal relational structure} of intermediate DiT representations to this frozen video teacher during training.

\noindent\textbf{Hidden-State Extraction}
We represent video latents as $z \in \mathbb{R}^{B \times C \times T \times H \times W}$, where $B$ denotes the batch size, $C$ the number of latent channels, $T$ the temporal length, and $(H,W)$ the spatial resolution. Let $(p_h,p_w)$ denote the DiT spatial patch size. Each temporal anchor is decomposed into $h_p w_p$ patches, where $h_p = H/p_h$ and $w_p = W/p_w$. We select the middle DiT blocks and extract their hidden states:
\begin{equation}
H^{(b)} \in \mathbb{R}^{B \times (T_f h_p w_p) \times d}
\end{equation}
where $T_f$ is the number of temporal anchors, and $d$ is the DiT hidden dimension. A lightweight MLP projector $\phi^{(b)}: \mathbb{R}^{d} \rightarrow \mathbb{R}^{D}$ maps hidden-state tokens to the teacher embedding space:
\begin{equation}
\hat{Y}^{(b)} = \phi^{(b)}\!\left(H^{(b)}\right) \in \mathbb{R}^{B \times (T_f h_p w_p) \times D}
\end{equation}

\noindent\textbf{Parameter-free Spatio-Temporal Grid Adaptation.} To capture fine-grained physical interactions, we must align the student and teacher representations on a unified dense grid. V-JEPA2 encodes the video into a feature volume $Z \in \mathbb{R}^{B \times t_g \times h_g \times w_g \times D}$, where $t_g, h_g, w_g$ denote the teacher's temporal and spatial grid dimensions. Unlike recent representation alignment methods (e.g., \citep{zhang2025videorepa}) that rely on parameterized convolutional downsamplers to match spatial dimensions—which can introduce structural biases and distort fine-grained geometric details—we employ a parameter-free \emph{continuous trilinear interpolation}. We reshape the projected DiT tokens $\hat{Y}^{(b)}$ into a 5D tensor and directly interpolate them to the teacher's grid size $(t_g, h_g, w_g)$. This naturally preserves the 3D continuous nature of physical space-time, yielding a synchronized student representation $\hat{Z}^{(b)} \in \mathbb{R}^{B \times t_g \times h_g \times w_g \times D}$.

\noindent\textbf{Physical Relation Alignment.}
Directly aligning absolute token values can be overly restrictive, potentially disrupting the generative prior of the VDM. Inspired by relational distillation methodologies \cite{zhang2025videorepa}, we propose to align the \emph{structural} knowledge of the physics world instead. While prior works focus on general semantic alignment, our objective explicitly targets kinematic constraints. We compute a joint spatio-temporal similarity matrix (Gram matrix) that simultaneously captures intra-frame spatial geometry and inter-frame causalities (e.g., how an object's position in frame $t$ relates to its surroundings in frame $t+k$).

Let $N_v = t_g h_g w_g$ denote the total number of spatio-temporal patch tokens in the video clip. We flatten the synchronized student and teacher features along the spatio-temporal dimensions, yielding $\mathbf{s}, \mathbf{t} \in \mathbb{R}^{B \times N_v \times D}$. 
For each batch, we compute the relation matrix $G \in \mathbb{R}^{N_v \times N_v}$, where each element represents the pairwise cosine similarity between two spatio-temporal tokens $i$ and $j$:
\begin{equation}
G_{i,j} = \frac{\left\langle s_i, s_j \right\rangle}{\|s_i\|_2 \|s_j\|_2}
\end{equation}
By computing this over the entire video volume, $G$ naturally encapsulates both spatial layouts and temporal motion trajectories. 

To transfer this physical prior, we minimize the discrepancy between the student's relation matrix $G^{(b)}_S$ and the teacher's relation matrix $G_T$. The physical alignment loss is defined using a margin-based L1 penalty:
\begin{equation}
L^{(b)}_{\text{Phys}} = \frac{1}{N_v^2} \sum_{i=1}^{N_v} \sum_{j=1}^{N_v} \max \left( 0,\ \left| \left[G^{(b)}_S\right]_{i,j} - \left[G_T\right]_{i,j} \right| - m \right)
\end{equation}
where $m$ is a margin parameter (e.g., $0.1$). The $\max(0, \cdot)$ operator introduces a tolerance threshold: it penalizes the model only when the relational discrepancy exceeds $m$. This prevents overly rigid feature matching, preserving the generative diversity of the diffusion model while still enforcing the necessary physical and kinematic constraints encoded by V-JEPA2.

\subsection{3D Geometry Injection}
\label{sec:3d_inject}
We opt to use depth, since it provides direct cues about 3D geometry, and is easy to obtain from the physics simulator. We add depth-feature alignment in latent space to inject physics clues into the video generation process.

\noindent\textbf{Depth Supervision for Physical Consistency.}
To ensure that the generated videos respect 3D scene geometry, we supervise depth prediction in the latent space. We modify the model to predict depth by attaching a lightweight 3D-convolution head. We start by extracting the intermediate hidden states from the middle layers of transformer blocks in DiT (\textit{e.g.}, blocks 12, 16, 20, 24). For each selected block $b$, the hidden tokens $H^{(b)}\in\mathbb{R}^{B\times N_{\text{seq}}\times d}$ are projected via a linear layer, reshaped to spatial feature maps,
and concatenated with the final video latents. Then the depth head predicts the depth latents $\hat{Z}^d\in\mathbb{R}^{B\times C\times T\times H\times W}$. Ground-truth depth maps are encoded as $Z^{d\star}$.


\noindent\textbf{3D Alignment Loss.}
To utilize multi-dimension depth as supervision to align the depth space and DiT feature space, we combine four complementary losses to supervise depth from both global and local perspectives:

\textbf{(1) Latent loss} enforces feature-space alignment, ensuring the depth representation is consistent with the video generation latent space:
\begin{equation}
\mathcal{L}_{\text{latent}}
= \frac{1}{BCTHW} \sum_{b,c,t,i,j}
\left\| \hat{Z}^d_{b,c,t,i,j} - Z^{d\star}_{b,c,t,i,j} \right\|_2^2.
\end{equation}
This loss provides strong supervision in the compressed latent space, improving training stability and convergence speed.

\textbf{(2) Pixel loss} operates on decoded depth maps to recover fine-grained details. We use a scale-and-shift-invariant (SI) metric~\cite{eigen2014depth} that is robust to global scale ambiguity inherent in monocular depth estimation:
%
\begin{equation}
\begin{aligned}
\mathcal{L}_{\text{pixel}}
&= \tfrac{1}{BT}\sum_{b=1}^{B}\sum_{t=1}^{T}
\min_{s_{b,t},\, t_{b,t}}\tfrac{1}{H_p W_p}
\sum_{i,j}\left\| s_{b,t}\,\hat{D}_{b,t,i,j} + t_{b,t} \right. \\
&\qquad\left. - D^{\star}_{b,t,i,j} \right\|_2^2
\end{aligned}
\end{equation}
where $\hat{D}, D^{\star}\in\mathbb{R}^{B\times 1\times T\times H_p\times W_p}$ are decoded depth predictions and targets at pixel resolution $(H_p, W_p)$, and $s_{b,t}, t_{b,t}$ are per-frame optimal scale and shift parameters computed via least-squares. By decoupling absolute depth values from relative structure, this loss focuses on geometric correctness rather than exact metric scale, which is crucial for generalization across diverse scenes.

\textbf{(3) Structure loss} preserves depth discontinuities and object boundaries via spatial gradient matching:
\begin{equation}
\begin{aligned}
\mathcal{L}_{\text{st}}
&=\frac{1}{BTH_pW_p} \sum_{b,t,i,j}
\Big(\left| \partial_x \hat{D}_{b,t,i,j} - \partial_x D^{\star}_{b,t,i,j} \right| \\
&\quad + \left| \partial_y \hat{D}_{b,t,i,j} - \partial_y D^{\star}_{b,t,i,j} \right|\Big)
\end{aligned}
\end{equation}
where $\partial_x \hat{D} = \hat{D}_{i,j+1} - \hat{D}_{i,j}$ and $\partial_y \hat{D} = \hat{D}_{i+1,j} - \hat{D}_{i,j}$ denote forward spatial differences in horizontal and vertical directions. This penalty encourages sharp depth transitions at object edges, preventing over-smoothing and improving the generation realism of details, particularly at occlusion boundaries.

\textbf{(4) Temporal loss} enforces motion consistency across frames, ensuring temporally coherent depth evolution:
\begin{equation}
\mathcal{L}_{\text{temp}}
= \frac{1}{B(T{-}1)H_p W_p}
\sum_{b=1}^{B} \sum_{t=1}^{T-1}
\big\|
\Delta_t \hat{D}_{b,t}
-
\Delta_t D^\star_{b,t}
\big\|_{1}
\end{equation}
where $\Delta_t D_{b,t} := D_{b,t+1} - D_{b,t}$ measures the frame-to-frame depth change at time $t$, capturing temporal dynamics. By matching predicted and target temporal gradients, this loss reduces flickering artifacts and enforces physically plausible motion flow, particularly important for rigid object translations and camera movements.

The combined weighted depth objective is then:
\begin{equation}
\mathcal{L}_{\text{3D}}
= \beta_\ell \mathcal{L}_{\text{latent}}
+ \beta_p \mathcal{L}_{\text{pixel}}
+ \beta_s \mathcal{L}_{\text{st}}
+ \beta_t \mathcal{L}_{\text{temp}}.
\end{equation}
This multi-faceted supervision provides complementary signals: latent alignment for global structure, pixel SI for metric-scale robustness, gradients for edge sharpness, and temporal consistency for motion smoothness. Together, they ensure geometrically accurate and temporally stable depth prediction, significantly improving the physical plausibility and 3D consistency of generated videos.

\subsection{Training Objective and Inference Behavior}
\label{sec:train_infer}

\noindent\textbf{Training Objective.} 
During the training phase, we fine-tune the base I2V diffusion transformer (e.g., via LoRA) using a combination of the standard generative objective and our proposed physical representation alignments. Since our foundation model Wan2.2 employs Flow Matching for generation, the primary reconstruction objective is the flow-matching loss $\mathcal{L}_{\text{FM}}$. To deeply internalize physical regularities, we jointly optimize this with the temporally-aware physical alignment loss ($\mathcal{L}_{\text{Phys}}$) and the multi-term 3D geometry alignment loss ($\mathcal{L}_{\text{3D}}$). The overall training objective is formulated as:
\begin{equation}
\mathcal{L} = \mathcal{L}_{\text{FM}} + \lambda_{\text{Phys}}\,\mathcal{L}_{\text{Phys}} + \lambda_{\text{3D}}\,\mathcal{L}_{\text{3D}}
\end{equation}
where $\lambda_{\text{Phys}}$ and $\lambda_{\text{3D}}$ are hyperparameters controlling the strength of the temporal-physical and spatial-geometric guidance, respectively.

\noindent\textbf{Inference.} At inference time, PhysAlign does \emph{not} require the video teacher nor the 3D/depth head. We discard all auxiliary branches and directly run the frozen I2V backbone with the learned LoRA adapters, using the same sampling scheme as the original flow-matching model. The model generates physically coherent videos conditioned solely on the input image and text prompt. Consequently, the runtime and memory footprint of PhysAlign are essentially identical to a standard Wan2.2+LoRA I2V generator, while benefiting from the physics-aware latent space shaped during training, making it highly efficient and practical for real-world video generation.

\section{Experiment}

We evaluate PhysAlign on standard image-to-video generation benchmarks and demonstrate that it substantially improves both physical coherence and overall video generation quality.

\vspace{-5pt}
\subsection{Implementation Details}

We build upon Wan2.2-I2V-14B~\citep{wan2025}, a state-of-the-art open-source DiT-based framework for image-to-video generation. Our model is fine-tuned using LoRA on the videos produced by our synthetic physics-aware data pipeline. We compare PhysAlign with strong recent open-source I2V baselines, including CogVideoX~\citep{yang2024cogvideox}, HunyuanVideo I2V~\citep{kong2024hunyuanvideo}, and Wan2.2~\citep{wan2025}.

\subsection{Evaluation Metrics}

Existing video generation benchmarks primarily emphasize semantic fidelity, often overlooking whether the generated motion adheres to basic physical laws. To comprehensively assess image-to-video performance, we evaluate our method using two complementary metric families.

\noindent\textbf{Physical Invariance Score (PIS).} Following~\citep{zhang2025morpheus, yuan2025newtongen}, we adopt the Physical Invariance Score to quantify the physical plausibility of object motion, while adapting the metric to suit the requirements of our task. PIS evaluates whether a trajectory preserves expected physical determinants $C$, computed case-by-case from velocity and acceleration statistics. We first apply SAM2~\citep{ravi2024sam2} to segment the moving object in each frame, extract its centroid sequence, and compute frame-wise velocities and accelerations. The final score is defined as:
\begin{equation}
\mathrm{PIS}=\left(1+\frac{C_\sigma}{|C_\mu|+\epsilon}\right)^{-1}.
\end{equation}

Although PhysAlign enforces physical constraints in 3D latent space, PIS is computed entirely in 2D image space. This aligns with human perception—physical plausibility is typically inferred from observed 2D motion. Under a pinhole camera, the 2D trajectory is a deterministic projection of the underlying 3D motion. Let $\Pi: (x,y,z)\rightarrow(u,v)$ denote the projection. The instantaneous 2D velocities and accelerations satisfy:
\begin{equation}
\begin{aligned}
\dot{u} &= \frac{f}{z}\Big(\dot{x}-\frac{x}{z}\dot{z}\Big), \qquad 
\ddot{u} = \frac{f}{z}\Big(\ddot{x}-\frac{x}{z}\ddot{z}\Big)
           -\frac{f}{z^2}\Big(\dot{z}\dot{x}-\frac{x}{z}\dot{z}^2\Big), \\
\dot{v} &= \frac{f}{z}\Big(\dot{y}-\frac{y}{z}\dot{z}\Big), \qquad 
\ddot{v} = \frac{f}{z}\Big(\ddot{y}-\frac{y}{z}\ddot{z}\Big)
           -\frac{f}{z^2}\Big(\dot{z}\dot{y}-\frac{y}{z}\dot{z}^2\Big),
\end{aligned}
\end{equation}
where $f$ is the focal length. Physically valid 3D motion yields smooth and sign-consistent 2D velocity and acceleration profiles, while non-physical motion (e.g., impulsive velocity jumps or inconsistent gravitational trends) produces detectable deviations.

For canonical projectile motion with initial speed $v$ and launch angle $\theta$, we have:
\[
v_x = v\cos\theta,\quad v_y = v\sin\theta - gt,\quad a_x = 0,\quad a_y = -g,
\]
These monotonicity and sign structures are preserved after projection up to a scale factor. Thus, acceleration smoothness and sign consistency in 2D space serve as a reliable proxy for validating 3D physical correctness.

In summary, PIS measures whether the projected motion exhibits physically consistent velocity and acceleration trends, enabling faithful evaluation of 3D physical validity using 2D observations.

\noindent\textbf{VBench I2V.}
V-Bench~\citep{huang2023vbench,huang2024vbench++} is a benchmark that evaluate the universal generation quality of the generated videos. We applied the VBench I2V evaluation benchmark which is specifically for the image-to-video generation quality. The evaluation dimension includes both I2V metrics metrics and quality metrics. For the I2V metrics, we mainly focus on the $i2v\_subject$  which measures the alignment between the subject in the input image and the subject in the resulting video, and $i2v\_background$, which assesses the coherence between the background scene in the input image and the generated video. We use VBench I2V on our physics-aware test set to evaluate general video generation quality, with a particular focus on motion quality.

\begin{figure*}[t!]
    \centering
    \includegraphics[width=\linewidth]{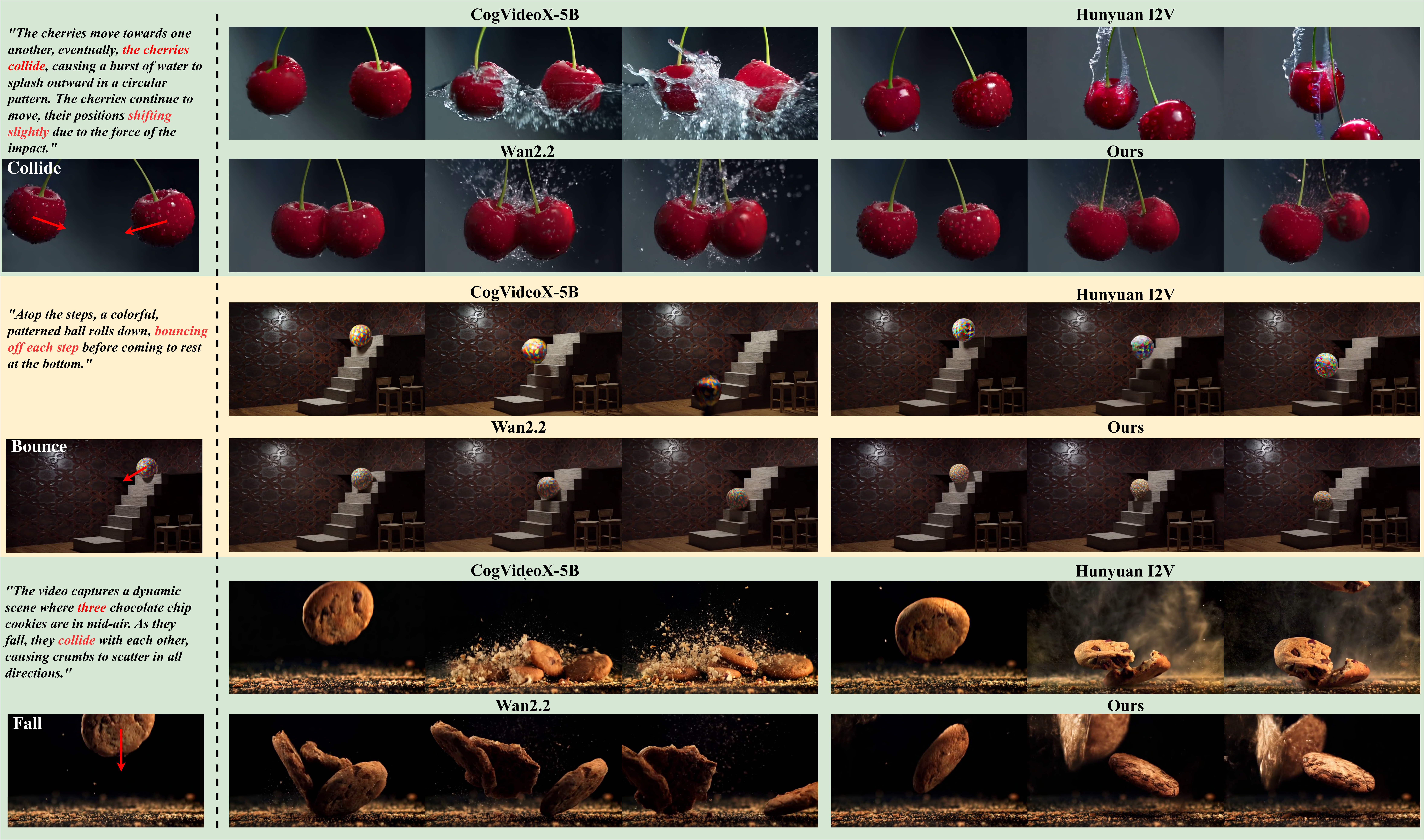}
    \caption{Comparison of our result with other baseline models on WISA-test set.  Results show that our method shows better understanding of the physics law, which demostrates our method's strong generalization ability to act as real-world simulator. Zoom-in for details. } 
    
    \label{fig:compare}
\end{figure*}

\begin{table}[t!]
\centering
\noindent
\vspace{0pt}
\centering
\setlength{\tabcolsep}{4pt}
\caption{Comparison of Physical Invariance Score (PIS, acceleration metrics) on Synthetic test set. Higher is better.}
\label{tab:pis_blender}
\scalebox{0.90}{%
\begin{tabular}{lccccc}
\toprule
\textbf{PIS} & \textbf{Ref} & \textbf{CogVideoX-5B} & \textbf{Hunyuan I2V} & \textbf{Wan2.2} & \textbf{Ours} \\
\midrule
$a_x$      & 0.701 & 0.350 & 0.571 & 0.520 & \textbf{0.632} \\
$a_y$      & 0.715 & 0.385 & 0.604 & 0.517 & \textbf{0.648} \\
$v_x$      & 0.790 & 0.494 & 0.704 & 0.679 & \textbf{0.746} \\
$v_y$      & 0.827 & 0.467 & 0.746 & 0.661 & \textbf{0.798} \\
$\Delta l$ & 0.777 & 0.227 & 0.592 & 0.546 & \textbf{0.641} \\
\bottomrule
\end{tabular}%
}
\end{table}

\hfill
\vspace{0pt}
\begin{table}[t!]
\centering
\setlength{\tabcolsep}{4pt}
\caption{Comparison of Physical Invariance Score (PIS, acceleration metrics) on WISA Test Set. Higher is better.}
\label{tab:pis_wisa}
\scalebox{0.90}{%
\begin{tabular}{lcccc}
\toprule
\textbf{PIS} & \textbf{CogVideoX-5B} & \textbf{HunyuanI2V} & \textbf{Wan2.2} & \textbf{Ours} \\
\midrule
$a_x$      & 0.444  & 0.568 & 0.444  & \textbf{0.604} \\
$a_y$      & 0.456  & 0.558 & 0.481  & \textbf{0.611} \\
$v_x$      & 0.678  & 0.708 & 0.669  & \textbf{0.775} \\
$v_y$      & 0.663  & 0.688 & 0.634  & \textbf{0.739} \\
$\Delta l$ & 0.372  & 0.416 & 0.317  & \textbf{0.451} \\
\bottomrule
\end{tabular}%
}

\end{table}


\begin{table*}[t!]
\centering
\caption{
VBench-I2V evaluation on the paired-subject set. Higher is better. 
\textbf{I2V-Metrics:} 
\textit{i2v\_subject} (subject consistency between input image and video), 
\textit{i2v\_background} (background consistency). 
\textbf{Quality Metrics:} 
\textit{subj\_consis.} (temporal subject consistency), 
\textit{bkg\_consis.} (temporal background consistency), 
\textit{motion\_smooth.} (motion smoothness), 
\textit{dynamic\_degree} (motion amplitude), 
\textit{aesthetic} (aesthetic quality), 
\textit{imaging} (imaging quality).
}
\label{tab:vbench_blender}
\scriptsize
\setlength{\tabcolsep}{2pt}  
\renewcommand{\arraystretch}{1.08}
\resizebox{\textwidth}{!}{%
\begin{tabular}{lcc|cccccc}
\toprule
\multirow{2}{*}{\textbf{Method}} &
\multicolumn{2}{c|}{\textbf{I2V-Metrics}} &
\multicolumn{6}{c}{\textbf{Quality Metrics}}\\
\cmidrule(lr){2-3}\cmidrule(lr){4-9}
& \textbf{i2v\_subject} & \textbf{i2v\_background}
& \textbf{subj\_consis.} & \textbf{bkg\_consis.}
& \textbf{motion\_smooth.} & \textbf{dynamic\_degree}
& \textbf{aesthetic} & \textbf{imaging} \\
\midrule
CogVideoX-5B & 0.816 & 0.870 & 0.772 & 0.876 & 0.985 & 0.540 & 0.406 & 0.636 \\
Hunyuan I2V           & 0.842 & 0.891 & 0.814 & 0.912 & 0.993 & 0.340 & 0.453 & 0.635 \\
Wan2.2 I2V            & 0.879 & 0.910 & 0.826 & 0.922 & 0.978 & 0.676 & 0.411 & 0.516 \\
\textbf{Ours}                  & \textbf{0.911} & \textbf{0.928} & \textbf{0.886} & \textbf{0.955} & \textbf{0.996} & \textbf{0.730} & \textbf{0.467} & \textbf{0.655} \\
\midrule
Reference Video       & 0.931 & 0.985 & 0.913 & 0.971 & 0.996 & 0.820 & 0.460 & 0.673 \\
\bottomrule
\end{tabular}
\!}%

\end{table*}

\begin{table*}[t!]
\centering
\caption{
VBench-I2V evaluation on the WISA-test set. Higher is better. 
}
\label{tab:vbench_wisa}
\scriptsize
\setlength{\tabcolsep}{2pt}  
\renewcommand{\arraystretch}{1.08}
\resizebox{\textwidth}{!}{%
\begin{tabular}{lcc|cccccc}
\toprule
\multirow{2}{*}{\textbf{Method}} &
\multicolumn{2}{c|}{\textbf{I2V-Metrics}} &
\multicolumn{6}{c}{\textbf{Quality Metrics}}\\
\cmidrule(lr){2-3}\cmidrule(lr){4-9}
& \textbf{i2v\_subject} & \textbf{i2v\_background}
& \textbf{subj\_consis.} & \textbf{bkg\_consis.}
& \textbf{motion\_smooth.} & \textbf{dynamic\_degree}
& \textbf{aesthetic} & \textbf{imaging} \\
\midrule
CogVideoX-5B & 0.953 & 0.953 & 0.918 & 0.941 & 0.988 & 0.337 & 0.454 & 0.618 \\
Hunyuan I2V           & 0.951 & 0.956 & 0.915 & 0.942 & 0.994 & 0.404 & 0.452 & 0.606 \\
Wan2.2 I2V            & 0.963 & 0.969 & 0.925 & 0.936 & 0.988 & 0.607 & 0.451 & 0.624\\
\textbf{Ours}                  & \textbf{0.975} & \textbf{0.981} & \textbf{0.934} & \textbf{0.951} & \textbf{0.995} & \textbf{0.635} & \textbf{0.458} & \textbf{0.643} \\
\bottomrule
\end{tabular}
\!}%
\end{table*}

\vspace{-10pt}
\subsection{Dataset}
\begin{figure}[t!]

    \centering
    \includegraphics[width=\linewidth]{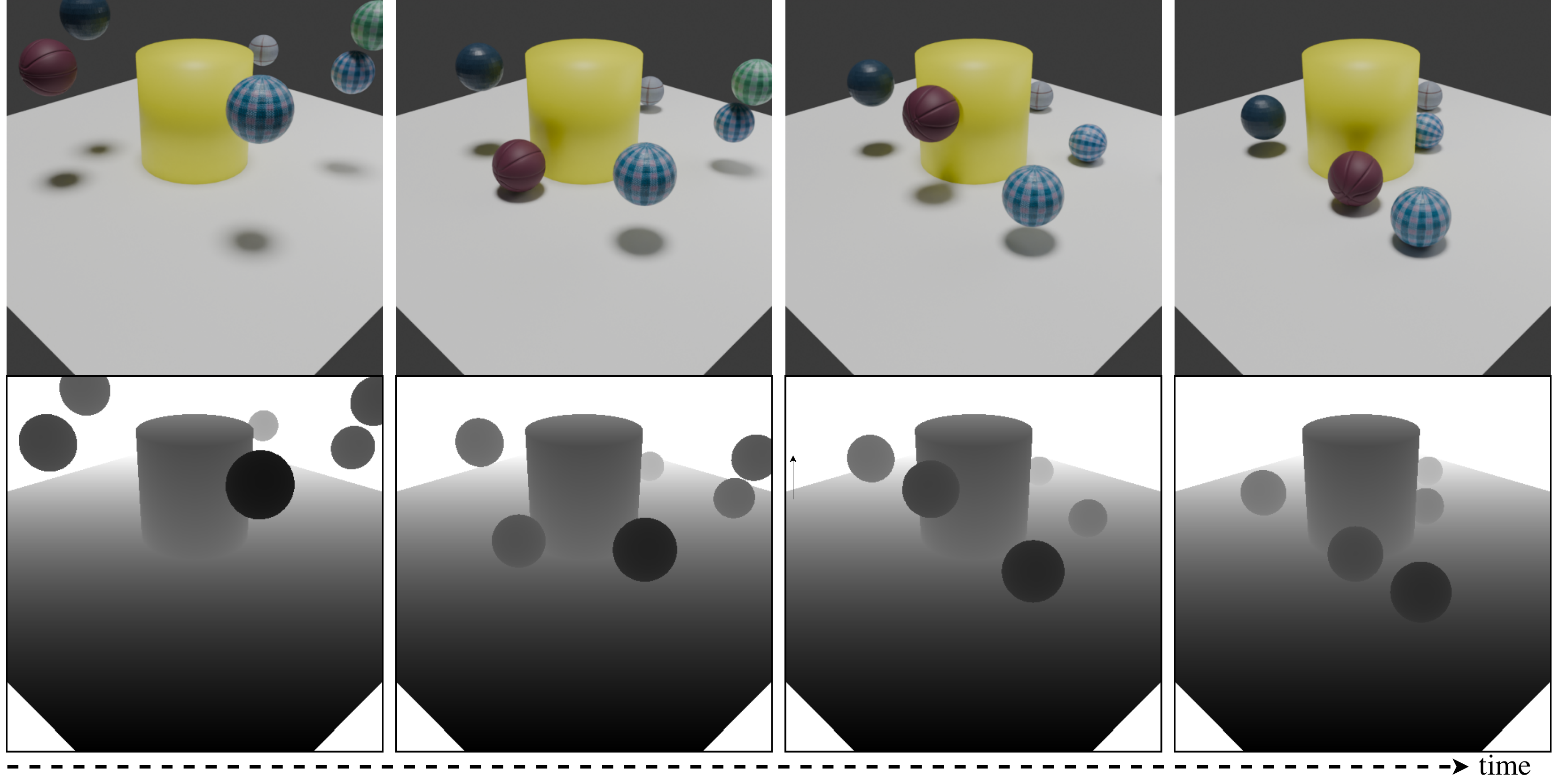}
    \caption{Visualization of the synthetic data generated though our data generation pipeline.}
    \label{fig:vis_data}
\end{figure}

\noindent\textbf{Synthetic data generation pipeline.} We show the detail of the data generation pipeline as below. The example generated data is shown in Fig~\ref{fig:vis_data}. We create 3k videos for training. The videos feature a range of objects, including balls, cylinders, cubes, cones, and icospheres, to better approximate real-world object distributions. 

Our generator samples the object type (balls with different categories such as basketball, soccer, tennis, bowling, as well as primitive rigid bodies like cubes, cylinders, cones, etc.) together with realistic masses and sizes, surface materials with varying friction, initial heights, applied force magnitudes and launch angles, the number of interacting objects, and optional lateral motion, so as to produce diverse scenes with rich collisions and occlusions. For each sampled configuration, we run Blender’s rigid-body simulator to render 90-frame sequences at 512×512 resolution, and export aligned RGB and depth along with detailed physics metadata (e.g., object mass, ball diameter, force in Newtons, starting height in meters, pixel-space coordinates of the applied force) and an automatically generated natural-language description. More visualization of our synthetic dataset in shown in the supplementary.

\noindent\textbf{WISA Dataset.}
WISA-80K~\citep{Wang_2025_Wisa} is a large-scale real video dataset containing explicit physics phenomena with rich text annotations, which is especially suitable for evaluating the physics law-following ability of the image-to-video generation models. We pick 200 videos that represent common physics laws and have a similar representation to our training data, including collision, rolling, bouncing, etc, extract the first frame as the reference image, and compose our WISA-Test set. We use this to evaluate the Syn-to-Real generalization ability of our proposed framework.

\subsection{Comparisons}

We present qualitative comparisons with recent state-of-the-art image-to-video models in Fig.~\ref{fig:teaser} and Fig.~\ref{fig:compare}. Across a wide range of motion types—such as rolling, bouncing, falling, and complex multi-object interactions—PhysAlign consistently produces trajectories that adhere to physically valid dynamics while preserving subject identity and aligning with the conditioning prompt. Competing approaches often suffer from identity drift, implausible accelerations, or inconsistent collision responses, whereas our method generates coherent motion that respects both the initial conditions and the underlying physical laws. Additional qualitative results and user study are provided in the supplementary material.

For quantitative comparisons, we evaluate PhysAlign using both the Physical Invariance Score (PIS) and VBench I2V metrics. On the Blender test set, results in Tab.~\ref{tab:pis_blender} and Tab.~\ref{tab:vbench_blender} show that PhysAlign achieves the highest performance across all PIS dimensions, including acceleration, velocity, and size-change consistency. These improvements highlight the effectiveness of physics-aware synthetic data in teaching the model to internalize accurate motion patterns.

We further evaluate cross-domain generalization on the WISA-Test set. As reported in Tab.~\ref{tab:pis_wisa} and Tab.~\ref{tab:vbench_wisa}, PhysAlign outperforms all baselines in both physics and perceptual metrics. The gains are especially pronounced for the \texttt{i2v\_subject} and \texttt{i2v\_background} metrics. This indicates that our synergistic alignment---intimately coupling explicit 3D geometry constraints with spatio-temporal relational embeddings---prevents structural distortion and substantially improves spatial fidelity with respect to the reference image. These results support our central claim: physical kinematics learned from abstract synthetic data, when deeply internalized via a unified representation alignment, transfer effectively to complex real-world scenarios.

We also report the evaluation results on PhysicsIQ~\citep{motamed2025physics} Benchmark, shown in Tab.~\ref{tab:physiciq}. Our method achieves consistent improvements over both the base model and the LoRA-trained variant on our synthetic dataset, demonstrating the effectiveness of our approach. We observe that most gains occur in the solid-mechanics category, which is consistent with our focus on rigid objects. We regard extending PhysAlign to more complex physical regimes and scenes as future work.

Overall, PhysAlign establishes a new state of the art in physics-aware image-to-video generation, excelling under both controlled synthetic evaluations and challenging real-world settings.

\begin{table}[t]
\centering
\footnotesize
\setlength{\tabcolsep}{3pt}
\renewcommand{\arraystretch}{0.6} 

\vspace{-8pt}
\caption{Comparison of \textbf{PhysicIQ score}. Higher is better.}
\vspace{-8pt}
\label{tab:physiciq}

\begin{tabular}{lcccc}
\toprule
\textbf{PhysicIQ (\%)} 
& CogVideoX-5B
& Wan2.2
& Wan2.2I2V+LoRA
& \textbf{Ours} \\
\midrule
Overall Score 
& 32.3
& 29.6
& 34.5
& \textbf{38.1} \\
\bottomrule
\end{tabular}
\end{table}

\begin{table}[t!]
\centering
\footnotesize
\setlength{\tabcolsep}{4pt}
\caption{Ablation Study of Physical Invariance Score (PIS) on Blender Test Set. Higher is better. FA: Feature Alignment, 3D-A: 3D Alignment.}
\label{tab:pis_ablation_blender}
\scalebox{1.0}{%
\begin{tabular}{lccccc}
\toprule
\textbf{PIS} & \textbf{Ref Video} & \textbf{Lora I2V} & \textbf{w/o FA} & \textbf{w/o 3D-A} & \textbf{Ours-Full} \\
\midrule
$a_x$ & 0.701 & 0.531 & 0.597 & 0.611 & \textbf{0.632} \\
$a_y$ & 0.715 & 0.548 & 0.602 & 0.615 & \textbf{0.648} \\
$v_x$ & 0.790 & 0.705 & 0.716 & 0.709 & \textbf{0.746} \\
$v_y$ & 0.827 & 0.689 & 0.768 & 0.762 & \textbf{0.798} \\
$\Delta l$ & 0.777 & 0.575 & 0.609 & 0.601 & \textbf{0.641} \\
\bottomrule
\end{tabular}%
}
\end{table}

\begin{table}[t!]
\centering
\footnotesize
\setlength{\tabcolsep}{4pt}
\caption{Ablation study of Physical Invariance Score (PIS) on the WISA-Test set. Higher is better. FA: Feature Alignment, 3D-A: 3D Alignment.}
\label{tab:pis_ablation_wisa}
\scalebox{1.0}{%
\begin{tabular}{lcccc}
\toprule
\textbf{PIS} & \textbf{LoRA} & \textbf{w/o FA.} & \textbf{w/o 3D-A.} & \textbf{Ours-Full} \\
\midrule
$a_x$      & 0.518 & 0.559 & 0.522 & \textbf{0.604} \\
$a_y$      & 0.505 & 0.578 & 0.549 & \textbf{0.611} \\
$v_x$      & 0.748 & 0.748 & 0.739 & \textbf{0.775} \\
$v_y$      & 0.669 & 0.679 & 0.662 & \textbf{0.739} \\
$\Delta l$ & 0.369 & 0.402 & 0.396 & \textbf{0.451} \\
\bottomrule
\end{tabular}%
}
\end{table}

\subsection{Ablation Study}

To better understand the contribution of each component in PhysAlign, we perform detailed ablation studies on both the Blender test set and the WISA-Test set, as summarized in Tab.~\ref{tab:pis_ablation_blender} and Tab.~\ref{tab:pis_ablation_wisa}. Removing the feature-level representation alignment results in clear degradation in motion coherence, including irregular velocity profiles, inconsistent acceleration patterns, and reduced stability of object identity over time. This indicates that semantic feature alignment provides crucial guidance for maintaining consistent latent motion representations.

Likewise, removing the 3D geometry alignment reduces depth-awareness and scale consistency, which manifests as incorrect size changes and weakened physical plausibility. Without this geometric supervision, the model struggles to preserve realistic perspective relationships and motion trajectories.

The full model, which integrates both semantic and geometric alignment, achieves the strongest performance across all physics-related and perceptual metrics. These ablations confirm that physics coherence in video generation arises from the complementary roles of feature-space alignment and 3D-space alignment, and that both are essential for achieving robust and physically consistent motion synthesis.

\section{Conclusion and Future Works}

We presented \textbf{PhysAlign}, an efficient adapter framework that bridges the gap between raw visual synthesis and rigid-body kinematics in Image-to-Video generation. Beyond achieving robust physical coherence, our work provides two critical insights. First, the remarkable data efficiency (requiring merely 3K clips) suggests that large video diffusion models already possess latent capacities for physical reasoning; PhysAlign's Gram-based spatio-temporal alignment and explicit 3D constraints act as a structural catalyst to unlock this potential. Second, our model's strong generalization from abstract synthetic shapes to complex real-world videos demonstrates that kinematic rules and visual appearance can be effectively decoupled within the latent space. Looking ahead, enhancing the model's interactive physical reasoning remains a vital direction. Future work will explore incorporating a physics agent that interacts with the generator via reinforcement learning, further advancing the paradigm of genuinely physics-grounded dynamic scene generation.

\balance
{
    \small
    \bibliographystyle{ieeenat_fullname}
    \bibliography{main}
}

\maketitlesupplementary

\appendix

\begin{table}[t]
\centering
\caption{User study results. We evaluate the user's preference from three aspects: Motion Physics Quality, Identity Physics Quality and Overall Video Quality. Higher is better.}
\small
\setlength{\tabcolsep}{2.5pt}  
\label{tab:user_study}
\begin{tabular}{lcccc}
\toprule
 & CogVideoX & Hunyuan & Wan2.2 & Ours \\
\midrule
Motion Physics Quality   & 1.671 & 1.543 & 2.557 & \textbf{3.236} \\
Identity Physics Quality & 1.757 & 1.679 & 2.536 & \textbf{3.343} \\
Overall Video Quality   & 1.771 & 1.607 & 2.643 & \textbf{3.286} \\
\bottomrule
\end{tabular}
\end{table}

\begin{table*}[t]
\centering
\caption{
Additional ablation study on VBench-I2V evaluation on the WISA-test set. Higher is better. }
\label{tab:vbench_wisa_supp}

\small
\setlength{\tabcolsep}{2pt}
\renewcommand{\arraystretch}{1.08}

\resizebox{\textwidth}{!}{%
\begin{tabular}{lcc|cccccc}
\toprule
\textbf{Method} &
\multicolumn{2}{c|}{\textbf{I2V-Metrics}} &
\multicolumn{6}{c}{\textbf{Quality Metrics}}\\
\cmidrule(lr){2-3}\cmidrule(lr){4-9}
& \textbf{i2v\_subject} & \textbf{i2v\_background}
& \textbf{subj\_consis.} & \textbf{bkg\_consis.}
& \textbf{motion\_smooth.} & \textbf{dynamic\_degree}
& \textbf{aesthetic} & \textbf{imaging} \\
\midrule
Train on WISA & 0.965 & 0.971 & 0.927 & 0.944 & 0.991 & 0.612 & 0.452 & 0.627 \\
DINOv3  & 0.972 & 0.978 & 0.928 & 0.944 & 0.993 & 0.617 & 0.455 & 0.633 \\
\textbf{Ours}  & \textbf{0.975} & \textbf{0.981} & \textbf{0.934} & \textbf{0.951} & \textbf{0.995} & \textbf{0.635} & \textbf{0.458} & \textbf{0.643} \\
\bottomrule
\end{tabular}
}
\end{table*}

\begin{table}[t]
\centering
\small
\setlength{\tabcolsep}{3pt} 
\caption{Additional ablation Physical Invariance Score(PIS) on WISA Test Set. Higher is better.}
\label{tab:pis_wisa_supp}
\begin{tabular}{lccccc}
\toprule
\textbf{PIS} & Train on WISA & DINOv3 & \textbf{Ours} \\
\midrule
$a_x$       & 0.538  & 0.592 & \textbf{0.604} \\
$a_y$       & 0.521  & 0.598 & \textbf{0.611} \\
$v_x$       & 0.753  & 0.764 & \textbf{0.775} \\
$v_y$       & 0.674  & 0.725 & \textbf{0.739} \\
$\Delta l$  & 0.399  & 0.436 & \textbf{0.451} \\
\bottomrule
\end{tabular}
\end{table}

\section{User Study}
We conduct a user study to compare our method against baseline approaches, as summarized in Tab.~\ref{tab:user_study}. Participants evaluate the generated videos from three perspectives: (1) \textit{Motion Physics Quality}, which measures how well the object motion adheres to physical laws; (2) \textit{Identity Physics Quality}, which reflects whether the subject’s deformation is consistent with physics; and (3) \textit{Overall Video Quality}, which captures general perceptual quality including visual fidelity and temporal consistency.

\begin{figure*}[t!]
    \centering
    \includegraphics[width=\linewidth]{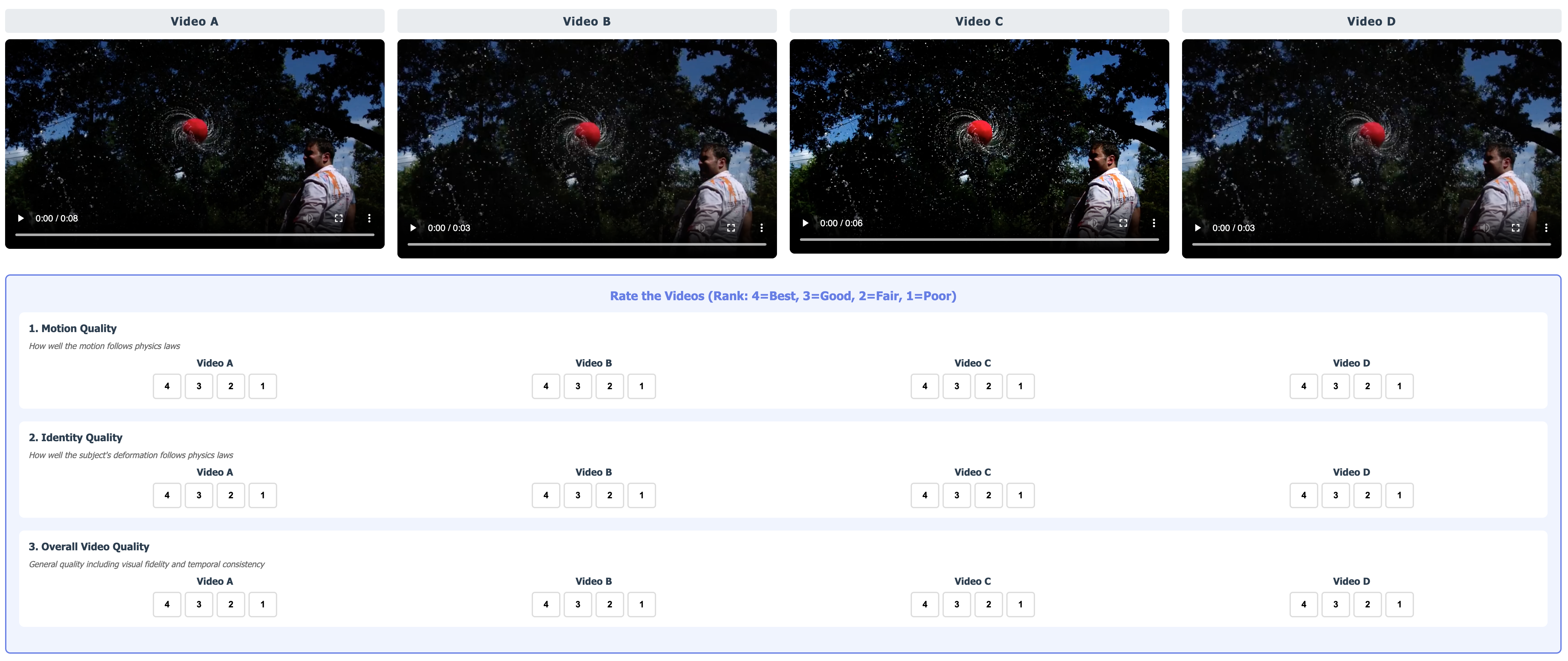}
    \caption{The visualization of our user study page.}
    \label{fig:user_study}
\end{figure*}

Each participant rates every video on a scale from 1 (poor) to 4 (best), and we report the average score for each method. The results show that our approach consistently receives higher human preference across all three aspects, demonstrating the effectiveness of the proposed physics-aware generation framework. We show the visualization of our user-study page in Fig.~\ref{fig:user_study}.

\section{Further Ablations}

\subsection{Comparison with Training on WISA}
We compare our model, trained on only 3k synthetic videos, with a LoRA-based I2V model trained directly on the real WISA dataset~\citep{Wang_2025_Wisa} to evaluate synthetic-to-real generalization. As shown in Tab.~\ref{tab:vbench_wisa_supp} and Tab.~\ref{tab:pis_wisa_supp}, despite using \emph{10$\times$ less} training data, our model achieves superior performance on both PIS and VBench I2V~\citep{huang2023vbench,huang2024vbench++}. These findings suggest that a compact synthetic dataset with accurate geometric and physical annotations can substantially reduce training cost while still improving physics-aware video generation quality.

From the comparison, we argue that the syn2real generalization is feasible because state-of-the-art pretrained video generation models already encode strong priors about visual dynamics. Our synthetic physic-clean data, enriched with explicit physics-aware annotations, acts as a targeted supervisory signal that elicits and reinforces these latent physical priors, enabling the model to better understand and utilize physics-based control cues. Our method can be trained on only 4 H100 GPUs than one day, offering an efficient and practical pathway for improving the physics-following ability of video generation models.

\subsection{Replacing the vision foundation model.}
We further perform an ablation by replacing the V-JEPA2 video teacher with DINOv3 while keeping the rest of the training pipeline unchanged. Although the DINOv3 variant still improves over training on WISA alone, it remains consistently inferior to our full model across both VBench I2V and PIS (Tab.~\ref{tab:vbench_wisa_supp}, Tab.~\ref{tab:pis_wisa_supp}). The performance gap is more pronounced on motion- and physics-related metrics, indicating that a video-native teacher provides more effective supervision for modeling temporal dynamics and physically coherent motion. We attribute this difference to the stronger spatio-temporal relational signals encoded by V-JEPA2, which better capture object evolution, motion continuity, and interaction patterns over time, whereas an image-centric teacher primarily emphasizes appearance and spatial semantics. These results further support the importance of temporally structured visual supervision for improving intuitive physics in image-to-video generation.

\section{Experiment Details}

We train our model on 4$\times$H100 GPUs, and the training converges in approximately 24  hours. 
The LoRA rank is set to 32 for all experiments. 
For a fair comparison with baseline methods, we use a video resolution of $480 \times 832$ during evaluation.
We use Wan2.2-14B model as our backbone. We compare our method with CogVideoX-5B, HunyuanVideo-I2V-13B and WAN-2.2-14B baseline.

\subsection{Training protocol.}
We train the high-noise and low-noise experts separately, using timestep ranges $[0,0.358]$ and $[0.358,1]$, respectively. LoRA adapters are inserted into the DiT modules \texttt{q}, \texttt{k}, \texttt{v}, \texttt{o}, and \texttt{ffn.0}/\texttt{ffn.2}. We optimize the model in \texttt{bfloat16} using AdamW with a constant learning rate of $1\times10^{-4}$ and weight decay $0.01$. 

\subsection{Input processing and sampling.}
Each training sample contains 49 frames, and the first frame is used as the conditioning image for I2V generation. Videos are resized by scaling to fill the target aspect ratio and then center-cropped. For physical feature alignment, we use a frozen V-JEPA2-ViT-L teacher with joint spatio-temporal relational distillation. We use the full clip for alignment, apply the loss at DiT block 16, set the alignment weight to 0.25 and the margin to 0.1, and resize teacher inputs to height 160 while preserving aspect ratio. Because V-JEPA2 uses a temporal tubelet size of 2, odd-length clips are truncated from 49 to 48 frames before teacher encoding. For the auxiliary depth branch, we use the multiscale head with 64 hidden channels and 3 layers. During evaluation, Wan-based methods are sampled with 49 frames, 50 inference steps.

\section{Additional Visualization Results}

\begin{figure*}[t!]
    \centering
    \includegraphics[width=\linewidth]{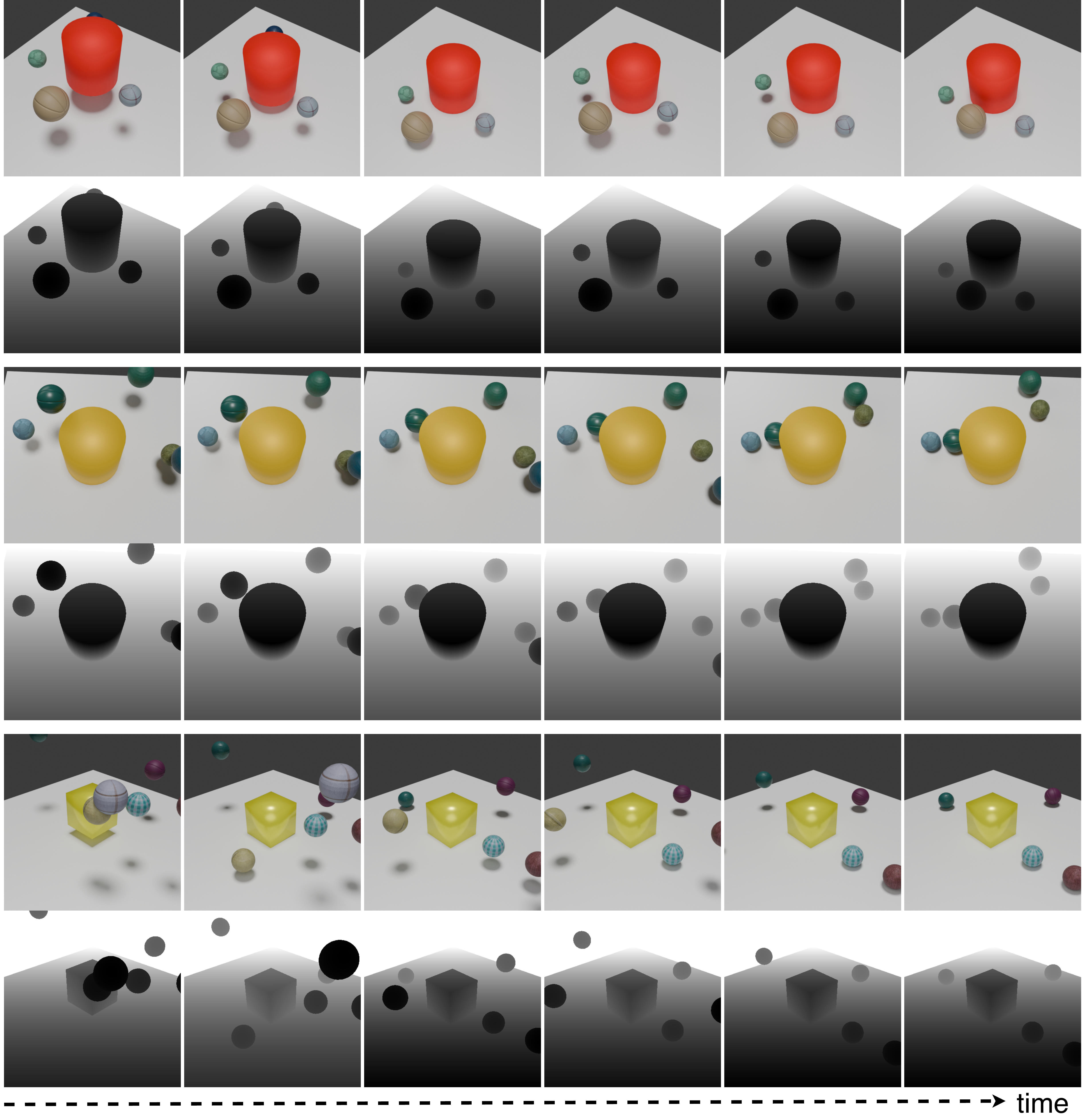}
    \caption{Additional visualizations of synthetic samples generated by our data construction pipeline.}
    \label{fig:data_visualization}
\end{figure*}

We also provide additional visualizations of our constructed synthetic data in Fig.~\ref{fig:data_visualization}.

\section{Impact Statement.}
This work aims to improve the physical plausibility of image-to-video generation through feature alignment and 3D representation alignment. Such improvements may benefit applications including video content creation, synthetic data generation, and visual world modeling, where more coherent object motion and scene dynamics are desirable. At the same time, we emphasize that our method is not intended as a substitute for precise physical simulation or scientifically rigorous modeling. Instead, PhysAlign focuses on \emph{intuitive physics}, i.e., motion patterns that appear more plausible to human observers and are better grounded in temporal and geometric structure. While our method improves motion realism and physical coherence, scenarios requiring more complex physical understanding or high-level reasoning remain challenging. We encourage the community to consider both the opportunities and limitations of such technology, especially in applications where stronger guarantees on physical accuracy are required.

\end{document}